\documentclass[letterpaper]{article} 
\usepackage{aaai24}  
\usepackage{times}  
\usepackage{helvet}  
\usepackage{courier}  
\usepackage[hyphens]{url}  
\usepackage{graphicx} 
\def\UrlFont{\rm}  
\usepackage{natbib}  
\usepackage{caption} 
\usepackage{algorithm}
\usepackage{algorithmic}

\usepackage{amssymb}
\usepackage{amsmath}
\usepackage{xcolor,tabularx,graphicx,amsmath,float,subfig}
\usepackage{booktabs}

\newcommand{\teal}[1]{\textcolor{teal}{{\bf [Teal: }{#1}{\bf ]}}}
\newcommand{\lucas}[1]{\textcolor{violet}{{\bf [Lucas: }{#1}{\bf ]}}}

\newcommand{\customparagraph}[1]{\vspace{0.1cm}\noindent\textbf{#1}\quad}

\usepackage{newfloat}
\usepackage{listings}
\DeclareCaptionStyle{ruled}{labelfont=normalfont,labelsep=colon,strut=off} 
\floatstyle{ruled}
\newfloat{listing}{tb}{lst}{}
\floatname{listing}{Listing}
\title{I Open at the Close:\\ A Deep Reinforcement Learning Evaluation of Open Streets Initiatives}
\author{
    R. Teal Witter\textsuperscript{\rm 1}\equalcontrib, Lucas Rosenblatt\textsuperscript{\rm 1}\equalcontrib
}
\affiliations{
    \textsuperscript{\rm 1}New York University\\

    \{rtealwitter, lucas.rosenblatt\}@nyu.edu
}

\usepackage{bibentry}

\begin{document}

\maketitle

\begin{abstract}
The open streets initiative ``opens'' streets to pedestrians and bicyclists by closing them to cars and trucks. The initiative, adopted by many cities across North America, increases community space in urban environments. But could open streets \textit{also} make cities safer and less congested? We study this question by framing the choice of which streets to open as a reinforcement learning problem. In order to simulate the impact of opening streets, we first compare models for predicting vehicle collisions given network and temporal data. We find that a recurrent graph neural network, leveraging the graph structure and the short-term temporal dependence of the data, gives the best predictive performance. Then, with the ability to simulate collisions and traffic, we frame a reinforcement learning problem to find which streets to open. We compare the streets in the NYC Open Streets program to those proposed by a Q-learning algorithm. We find that the streets proposed by the Q-learning algorithm have reliably better outcomes, while streets in the program have similar outcomes to randomly selected streets. We present our work as a step toward principally choosing which streets to open for safer and less congested cities. All our code and data are available on Github.\footnote{\url{https://github.com/rtealwitter/OpenStreets}}

\end{abstract}

\section{Introduction}
Traffic congestion is at its best inconvenient and at its worst very dangerous. 
In 2022, American drivers spent an average of 51 hours
stuck in traffic, representing an estimated
81 billion dollars in productivity loss \cite{inrix2022global}.
Furthermore, a by-product of increased traffic in urban areas is an \textit{increase in traffic collisions }\cite{retallack2019current}. Both traffic and collisions can be mitigated by intelligent
road network design, but urban road networks are already built
and new infrastructure projects in cities
can be prohibitively expensive \cite{siemiatycki2015cost}.

One potential solution is to ``open\footnote{In this work, we will use ``open'' to mean  \textit{closing} a street to vehicular traffic and \textit{opening} a street to pedestrians and cyclists.}'' existing roads to pedestrians and bicyclists by closing them to cars and trucks
\cite{kuhlberg2014open, bertolini2020streets}.
Generally, open streets initiatives provide a communal space for people living in urban environments.
As expected, the initiatives have positive impacts, including on the physical health of the participants \cite{cohen2016ciclavia, sharples2014needs}.
But there is some evidence that opening streets also improves traffic and safety.
Two prominent examples are Times Square and Herald Square in Manhattan, NYC, which were turned into pedestrian plazas in 2009.
After the squares opened, there was a reduction of approximately 15$\%$ travel times for routes along Broadway and a 63$\%$ reduction in injuries to drivers and passengers on the avenue \cite{grynbaum2010new}.

Numerous exogenous variables make an empirical analysis of the effects of opening streets challenging.
As a result, cities generally use a proposal process to identify which streets to open.
Identifying candidate streets for opening has been studied \cite{youn2008price, rhoads2021sustainable}; however, previous work does not simulate the effects of proposed open streets nor does it systematically consider the impact of open streets on public safety and vehicular congestion. In this work, we do both.

\begin{figure*}[h!]
  \centering \includegraphics[width=0.85\textwidth]{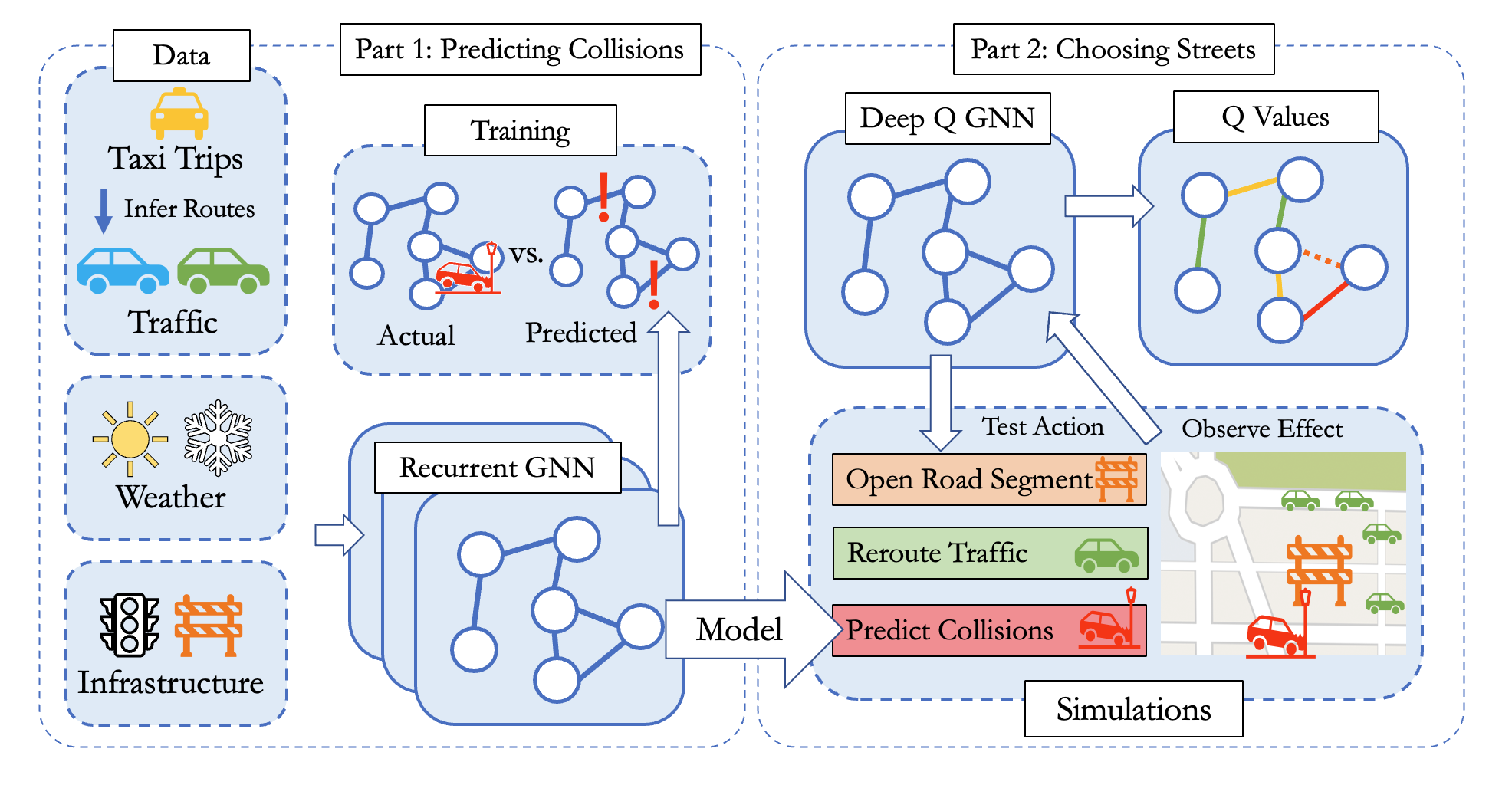}
  \caption{In Part 1 of our work, we build
  a recurrent Graph Neural Network (GNN) to predict
  collisions. In Part 2, we train a deep Q GNN to
  reduce traffic and collisions by opening
  road segments
  (using our Part 1 model to measure collisions after hypothetical road openings.)
  Our deep GNN's Q-values represent the expected long-term reduction in traffic and collisions of opening a road.} 
  \label{fig:summary}
\end{figure*}

\customparagraph{Contributions}
In the first part of our work, we build an improved model for predicting collisions, evaluated on granular and comprehensive data. 
We consider a wider time frame (days, months and years) and a larger space (the entirety of Manhattan) for predicting collisions than prior work.
We are the first to: (1) use years of data that account for seasonal and annual variations in traffic and exogenous variables like weather (prior work only used several months), (2) use all negative and positive examples (prior work subsampled to enforce class balance), and (3) take a global view of the road network, predicting collisions at the city level while taking into account local information (prior work used only a few block radius around a collision). 
We compare several models for the prediction task.
Our best model uses recurrent layers to capture \textit{short-term} temporal dependencies and graph convolutional layers to capture \textit{spatial} dependencies of our data.
Finally, we analyze the importance of features in the best model and discuss connections to prior work in the transportation literature.

In the second part of our work, we use a deep learning approach to evaluating the efficacy of the NYC Open Streets program.
To the best of our knowledge, we are the first to formulate the problem of opening roads in the language of reinforcement learning.
We simulate road openings in real historical days.
For each simulated day, we estimate traffic as the total car density per capacity of each road and estimate collision risk using our collision prediction model from the first part of our work. 
We train a deep Q-learning model to output the long term value of opening each road segment.
We find that the streets opened by the NYC Open Streets program have similar performance to randomly selected streets and that the streets in the program are geographically concentrated in certain Manhattan neighborhoods.
In contrast, the streets with the highest Q-values consistently reduce collisions and traffic, with a more equal distribution across Manhattan.
As a result, we recommend using our Q-learning model as an additional method for evaluating streets in the NYC Open Streets program.

Figure \ref{fig:summary} summarizes the two parts of our work.






\subsection{Related Work}

\customparagraph{Open Streets}\label{subsec:openstreets}
Open street initiatives are multifaceted in function:
they allow more space for people to safely exercise and traverse the city, they promote a decrease in vehicle traffic and carbon emissions, and they provide expanded outdoor space for businesses, particularly restaurants \cite{hazarika2021reclaiming}.
However, the initiatives have faced some challenges. Critics have pointed out a lack of equity in the implementation, arguing that more affluent neighborhoods have benefited disproportionately from the program in part because streets are chosen through a proposal process \cite{hazarika2021reclaiming}.
Prior work has used a coarse approximation of a city to suggest which roads to open \cite{youn2008price} and suggested roads based on side walk infrastructure \cite{rhoads2021sustainable}.
But we are not aware of any work that simulates the effects of opening streets or systematically considers the impact on safety.

\customparagraph{Collision Prediction} 
Many papers have used traditional machine learning techniques to solve the problem
of predicting collisions
\cite{Cheng2019NonlinearData, Baloch2020MachineSeverity,Auret2012InterpretationForests}.
However, there are complicated spatial and temporal dependencies in road networks
and collision dynamics.
To fit these non-linear patterns, there has been substantial
interest in deep learning techniques to predict collisions
\cite{Lin2021IntelligentApproach}.
In \citet{Ren2017APrediction}, they use the recurrent long-short-term memory architecture
to capture temporal relationships.
In \citet{Zhao2019ResearchVANET}, they use standard convolutional layers to capture
both temporal and spatial relationships.
Our work is most similar to \citet{yu2021deep}.
They use taxi, weather, road infrastructure, and point data
to predict collisions with a deep graph neural network which captures
both temporal and spatial relationships.
However, they only consider a two-month period and down-sample the number of events to equalize the number of collisions and non-collisions, reducing data-scale drastically.

\customparagraph{Traffic Prediction}
The success of our work is predicated on inferring traffic flow from available taxi trip data as accurately as possible.
\citet{yu2021deep} focused on the prediction of collisions in Beijing, where data on the exact location of the entire Beijing taxi fleet is available in increments of 5 minutes.
However, in our setting in Manhattan, we do not know the exact location of taxis during their trips. That said, we \textit{do} have access to trip start and end GPS coordinates for the period from 2013 to 2015, and we rely on this data for traffic inference \cite{nyctaxidata}.\footnote{Coordinates have not been reported since 2015 due to privacy concerns.} We note that the
NYC taxi trip data before 2016 have been successfully used in a variety of applications
such as fleet dispatching and routing
\cite{Bertsimas2019OnlineApplications, Deri2016BigCity},
taxi supply and demand
\cite{Yang2017ModelingData},
traffic safety \cite{Xie2017AnalysisHotspots},
and predicting congestion
\cite{Zhang2017AnalyzingData}.
Machine learning techniques have also been quite popular for supervised
learning tasks related to taxi data.
These works include identifying
areas of interest \cite{liu2021identifying},
traffic \cite{Yao2018DeepPrediction, wu2016interpreting, zhang2021graph},
taxi demand \cite{Luo2022Spatial-TemporalForecasting},
speeding drivers \cite{zhong2022taxi}, 
payment type \cite{Ismaeil2022UsingTaxi}, 
and classifying collisions \cite{Abeyratne2020ApplyingCity, Bao2021ExploringData}.

\section{Background}

Below we summarize the notation and ideas underlying
our use of graph neural networks and reinforcement learning.

\customparagraph{Graph Neural Networks}\label{subsec:gnn}
Graph Neural Networks (GNNs) are a popular
choice for exploiting the structure in graphs 
\cite{welling2016semi}.
A graph consists of a set of nodes $\mathcal{V}$
and a set of directed edges $\mathcal{E}$.
In our work, the nodes represent the segments of the road and the edges represent the intersections that connect them.

Formally, the local structure is captured in a GNN
through graph convolutional layers.
Consider a node $v \in \mathcal{V}$ and its
representation $\mathbf{x}^\ell_v \in \mathbb{R}^d$
at the $\ell$th layer of a GNN.
Then we build its representation $\mathbf{x}^{\ell+1}_v 
\in \mathbb{R}^{d'}$ at the next layer
by applying a graph convolution with parameters
$\mathbf{\Theta} \in \mathbb{R}^{d' \times d}$.
In particular,
\begin{align*}
\mathbf{x}^{\ell+1}_v = \sigma \left( \mathbf{\Theta}
\sum_{u: (u,v) \in \mathcal{E}}
\mathbf{x}^\ell_v w_{u,v} \right)
\end{align*}
where $w_{u,v}$ is the normalized weight
of the edge between $u$ and $v$
and $\sigma$ is an activation function.

An advantage of GNNs in our setting 
is that GNNs learn weights that can be applied to exploit
connections in \textit{any} graph, provided
the node features stay consistent.
The size of the graph changes in  our reinforcement learning problem since each action removes a road segment.
Furthermore, a GNN can capture short term temporal information by passing in a hidden state to each layer, which is important in our setting as short term temporal information is more important than long term. 
This architecture is called a `recurrent GNN' (RGNN) \cite{seo2018structured}. 

\begin{figure}[h!]
    \centering \includegraphics[width=0.8\linewidth]{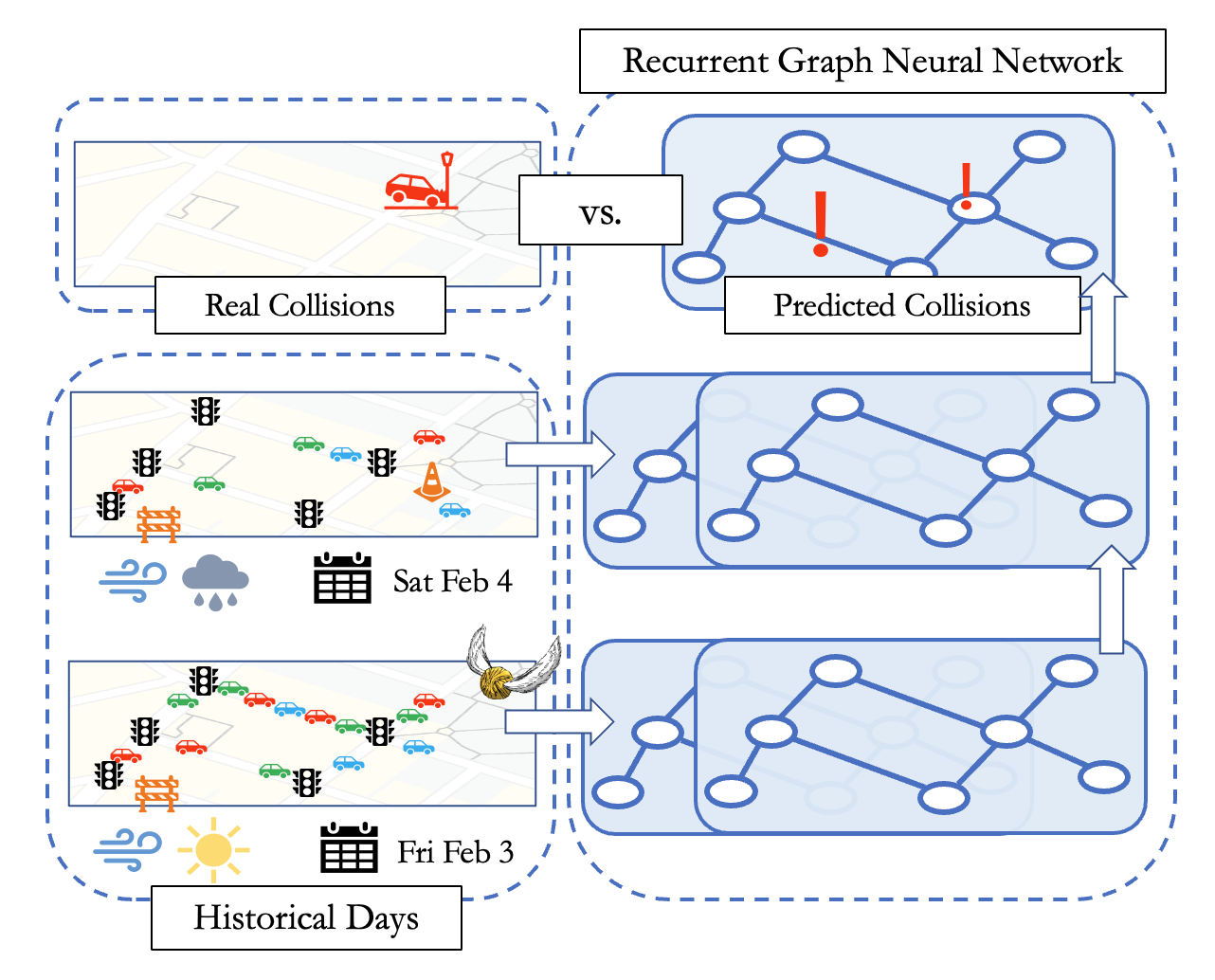}
    \caption{Using traffic, weather, and infrastructure
    data, we train a recurrent GNN to predict collisions.
    The recurrent connections capture short term temporal
    dependencies like weather.
    We use weighted cross entropy loss to compare
    the real and predicted collisions.}
    \label{fig:recurrent_gnn}
\end{figure}

Figure \ref{fig:recurrent_gnn} describes how the RGNN captures the spatial and temporal relationships in the collision prediction problem.

\customparagraph{Reinforcement Learning}\label{subsec:rl}
Reinforcement Learning (RL) is a collection of techniques
for optimization in online learning settings
\cite{sutton2018reinforcement, moerland2023model}.
In RL, a stochastic agent makes decisions in an environment and receives feedback, in turn learning a good policy.

Formally, we say the state space $\mathcal{S}$ 
is the set of all possible realizations of an environment
while the action space $\mathcal{A}$
is the set of all possible options we can take in
an environment.
An agent is presented with a state and
must then choose an action to take.
Once an action is chosen, the agent transitions
to a new state using a stochastic transition function
$f: \mathcal{S} \times \mathcal{A} \rightarrow \mathcal{S}$.
The agent also receives a real-valued reward
for taking an action in a given state 
according to a stochastic reward function
$r: \mathcal{S} \times \mathcal{A} \rightarrow \mathbb{R}$.
A ``good'' policy is then one that maximizes 
the reward an agent receives in expectation.

In this work, we consider Q-learning because it is generally
more sample efficient than other techniques and produces useful intermediate values \cite{jin2018q}.
The $Q$ function, the namesake of Q-learning,
is used to find and exploit
states that produce a high reward.
Consider a stochastic sequence
of states and actions $(s_0, a_0, s_1, a_1, \ldots)$
where $s_{i+1}=f(s_i, a_i)$ and actions are selected
according to our policy.
Then we can write
\begin{align*}
Q(s,a) = \mathbb{E}\left[ 
\sum_{i=0}^\infty r(s_i, a_i) \gamma^i
\right]
\end{align*}
where $0 < \gamma < 1$ is some discount factor
chosen so that we focus on near term reward.
If we had these $Q$ values, then our policy
should choose the next action in state
$s$ by calculating $\arg \max_a Q(s,a)$.

\begin{table*}[t]
\centering
\caption{Results of collision prediction models. Overall support in the test set was 1,803,363 observations: 1,789,838 negative and 13,525 positive examples. The $\pm$ denotes standard deviation 10 random seeds. Since the F1-score ignores the imbalanced nature of our data, we use the macro average recall to select the best model.}
\begin{tabular}{lrrrrr}
\toprule
 Model & F1-score & Recall  (Negative) &  Recall (Positive) &\textbf{Recall  (Macro Average)}  \\
\midrule 
Gaussian NB                    & 0.97$\pm$0.0001 & 0.95$\pm$0.0001 & 0.15$\pm$0.0001 & 0.55$\pm$0.0001 \\
LightGBM                      & 0.78$\pm$0.0005 & 0.64$\pm$0.0006 & 0.80$\pm$0.0003 & 0.72$\pm$0.0002 \\
XGBoost                       & 0.80$\pm$0.0001 & 0.67$\pm$0.0001 & 0.81$\pm$0.0001 & 0.74$\pm$0.0001 \\
DSTGCN \cite{yu2021deep}      & 0.67$\pm$0.2600 & 0.56$\pm$0.2701 & 0.59$\pm$0.1070 & 0.57$\pm$0.0401 \\
Graph WaveNet \cite{wu2019graph}& 0.75$\pm$0.0121 & 0.61$\pm$0.0160 & 0.68$\pm$0.0006 & 0.64$\pm$0.0080 \\
Recurrent GNN (Lite) & 0.86$\pm$0.0130 & 0.77$\pm$0.0200 & 0.68$\pm$0.0215 & 0.73$\pm$0.0043 \\
\textbf{Recurrent GNN }       & 0.87$\pm$0.0064 & 0.78$\pm$0.0102 & 0.74$\pm$0.0157 &\textbf{ 0.76$\pm$0.0040} \\
\bottomrule
\end{tabular}
\label{tab:collision_results}
\vspace{-0.3cm}
\end{table*}

This observation motivates the Bellman equation, a
natural criteria for our $Q$ function,  where $s' = f(s,a)$:
\begin{align*}
Q(s,a) = r(s,a) + \gamma \max_{a'} Q(s', a').
\end{align*}
In our setting, we will use a neural network
for the $Q$ function with parameters $\theta$.
Then the loss function is given by
\begin{align*}
\mathcal{L}(\theta) = \left(r(s, a) 
+ \gamma \max_{a'} Q(s', a') - Q(s,a)\right)^2.
\end{align*}
Notice that the $Q$ function appears in two places.
During optimization, we freeze weights of
the `target' $Q$ function on the left and update
the $Q$ function on the right. Q-learning allows us to efficiently
train the $Q$ function; notice that we only
need the 4-tuple $(s,a,s',r)$ to compute the loss.

\section{Part 1: Predicting Collisions}\label{sec:collisions}


In this section, we describe our work on predicting collisions. Compared to previous work, our work considers a longer time frame (three years instead of several months) and includes all collision \textit{and} non-collision events available in our data set (instead of down-sampling non-collisons). In this setting, we find that our recurrent GNN outperforms the state-of-the-art models from prior work.

\customparagraph{Data}
In order to predict collisions, we leverage infrastructure data
like road attributes, day-specific weather conditions, and traffic data.
The data we use are granular: we have information for all road segments in Manhattan every day over a three-year period.
Each road segment is defined as the portion of a street between two intersections, yielding 19,391 segments in Manhattan.
Unfortunately, traffic data is not available at our geographic and temporal scale.
Instead, we infer overall traffic in our road network using a massive set of start and end locations from taxi trips.
We use Dijkstra's shortest-path algorithm to efficiently calculate where taxis, and we assume other vehicles, likely drove.
Since our data set is massive (10 to 15 million taxi trips per month), we used simplified shortest paths and local rerouting.

\customparagraph{Problem} We formulate collision prediction as a binary classification problem: ``Did a collision occur at this road segment on this day?'' Note that we could also frame the problem as ``How many collisions occurred at this road segment on this day?'' However, due to the sparsity of collisions, we chose the more tractable binary classification framing over the regression framing.

\customparagraph{Imbalanced Classification}
Collisions over a road network are sparse. In our training data, we had more than 21 million total observations but only 160,549 collision observations (each observation is a road segment on a particular day). So, as a fraction of the data, only $0.76\%$ of our observations came from the positive class. 
There are several standard approaches to imbalanced classification problems. Perhaps the most common approach is to down-sample the majority class, equalizing the total number of negative and positive examples.
This is the approach taken by prior work on collision prediction \cite{yu2021deep, Lin2021IntelligentApproach}.
Unfortunately, down-sampling requires throwing away
over $99\%$ of our data, which is also a major drawback of the prior work. 
We found that the second natural approach, up-sampling, was inappropriate for our task
because of the high variance of collisions and
difficulty in characterizing the underlying distribution (which is essentially our learning problem).
A third approach, and the one that we implement, 
is to weight the loss functions of our collision 
prediction models so that the positive examples
have the same importance as the negative examples. 
The benefit of this approach is that we can utilize all 
our data for learning without risking the introduction 
of noisy synthetic data.

\customparagraph{Metrics}
Overall accuracy is not a helpful metric since a model that always predicts ``no collisions'' would have over 99\% accuracy. Instead, we seek a high \textit{recall} model for both the positive \textit{and} negative class, to prioritize the prediction of possible collisions. 
However, a model that over-predicts collisions, thus drowning out useful signal for our downstream RL, is also undesirable. 
Therefore, the \textit{unweighted} average of recall between the positive and negative class (also known as \textit{macro average recall}) is our preferred metric; we find that this metric reflects our focus on recall for the positive class while not overly discounting recall for the negative class.

\customparagraph{Models}
We started with standard machine learning models like Logistic Regression, Random Forest, and Gaussian Naive Bayes classifiers.
Among these, only the Gaussian Naive Bayes classifier demonstrated non-trivial recall for the positive class.
We then considered boosting algorithms XGBoost and LightGBM.
Both models were strong baselines likely because of their known effectiveness on data sets like ours with a large number of features \cite{borisov2022deep}.

We then evaluated DSTGCN which was specifically designed for the collision prediction task \cite{yu2021deep}.
We found DSTGCN performed poorly, perhaps because of the much larger scale of our data and problem.
We also evaluated two widely successful architectures for traffic prediction: Graph WaveNet \cite{wu2019graph} and DCRNN \cite{li2017diffusion}.
Unfortunately, DCRNN is too slow and memory intensive for our problem's scale.\footnote{
For its original task on 207 road segments with four months of data, training takes days to run even on an Nvidia A100. Our problem has more than 90x the number of road segments and 9x the number of samples. Even though we tried, training the DCRNN took too long on our data.}

We next evaluated a recurrent GNN (RGNN).
We hypothesized that the road structure and traffic patterns interact temporally \textit{in the short term} and that the recurrent layers could successfully capture these relationships.
Unlike Graph WaveNet, the RGNN uses a \textit{fixed graph structure} which we hypothesize enables it to achieve higher performance on the large network we consider.

Table~\ref{tab:collision_results} summarizes our findings.
Each model was hyperparameter tuned and, when applicable (i.e. in the deep learning setting), trained for 100 iterations. 
We report on the average performance (plus or minus the standard deviation) over ten random initializations of each model on the same train and test sets.

\begin{figure}[ht!]
    \centering \includegraphics[width=.95\linewidth]{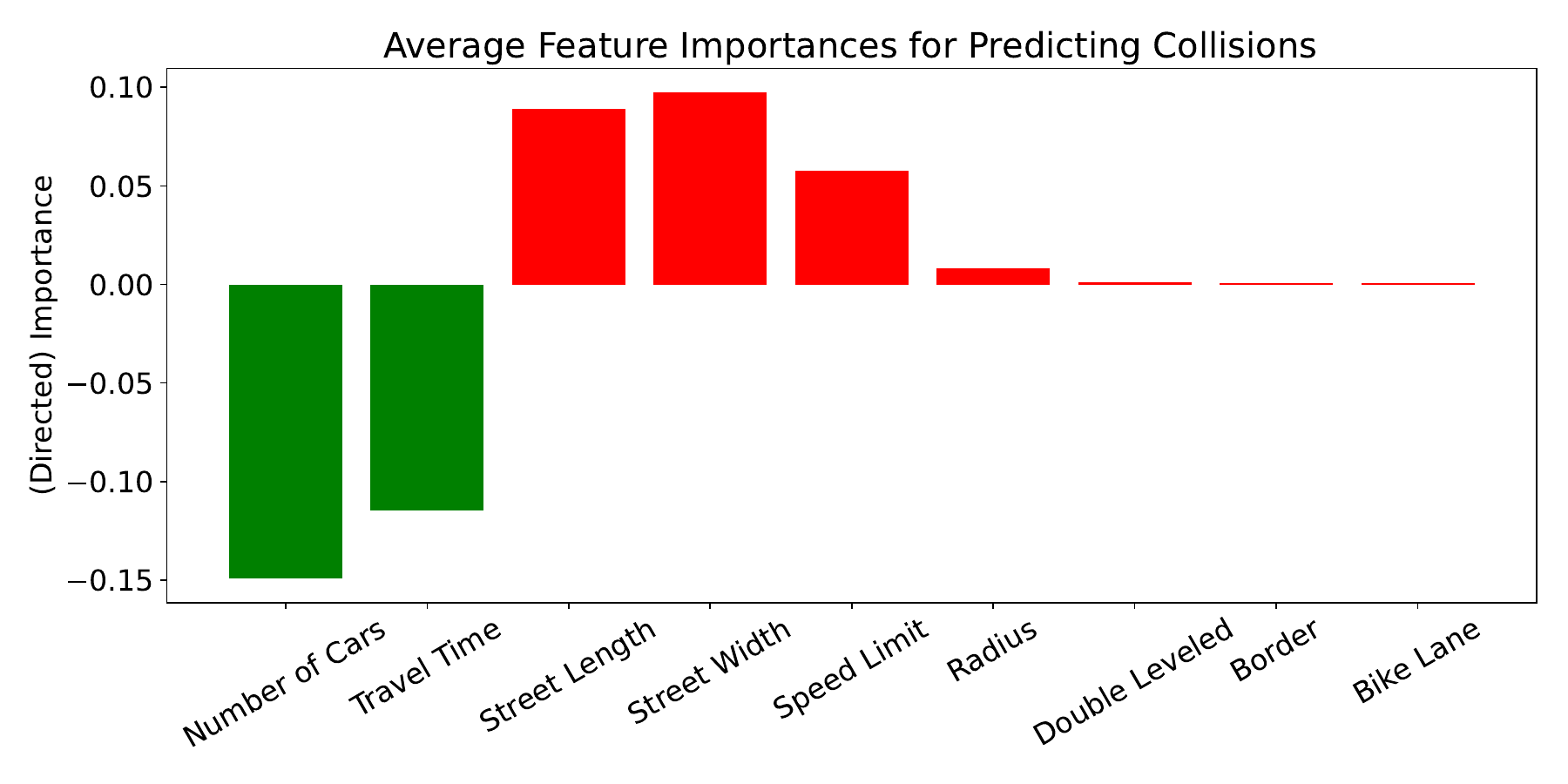}
    \caption{We use the integrated gradients method to compute the feature importance of the trained RGNN \cite{sundararajan2017axiomatic}.
    Features with negative importance (green) are associated with decreased collision risk while
    features with positive importance (red) are associated with increased collision risk.}
    \label{fig:collision_importance}
\end{figure}

Figure \ref{fig:collision_importance} plots the average predictive effect of the most important features.
The number of cars on a road segment and the travel time (in ideal conditions) generally reduce the risk of collisions.
We hypothesize this is because driving is cognitively easier in slower conditions \cite{nilsson2017vehicle}.
The remaining features we plot all generally increase the risk of collisions.
Street length, street width, and speed limit are all associated with higher speed roads which make collisions more likely \cite{das2023impact}.
The features with the next largest effects---the radius of curved roads, double level roads, and roads on the border of Manhattan---are all associated with the highways into and out of the island.
It is well-documented that locations with speed variation are correlated with more collisions \cite{li2013evaluation}.
Finally, bike lanes also have a (small) effect on collisions perhaps because of the dangers of biking in Manhattan \cite{chen2012evaluating}.

\section{Part 2: Choosing Streets}
In this section, we consider the problem of
choosing which road segments to open as a mechanism for 
reducing traffic and collisions.
A natural idea is to choose the road segments with
the highest levels of traffic and most collisions.
However, there are two issues with this approach:
\begin{enumerate}
    \item There are complicated endogenous effects. For example, if they are rerouted from road segments with traffic, cars may clog smaller streets or exacerbate gridlock in other heavily congested areas. In addition, more complicated traffic patterns that increase cognitive load can raise the risk of collisions \cite{engstrom2017effects}.
    \item There are complicated temporal dynamics so road segments that are beneficial to open in some conditions may be quite harmful to open in others. Days of the week, weather patterns, and special events all impact where and how people drive.
\end{enumerate}

We address these two issues by framing the problem
of opening road segments as a reinforcement learning problem.
In particular,
we incorporate temporal dynamics by considering sequences
of days and we incorporate endogenous effects by propagating
road openings through time.

\begin{figure}[ht!]
    \centering \includegraphics[width=1\linewidth]{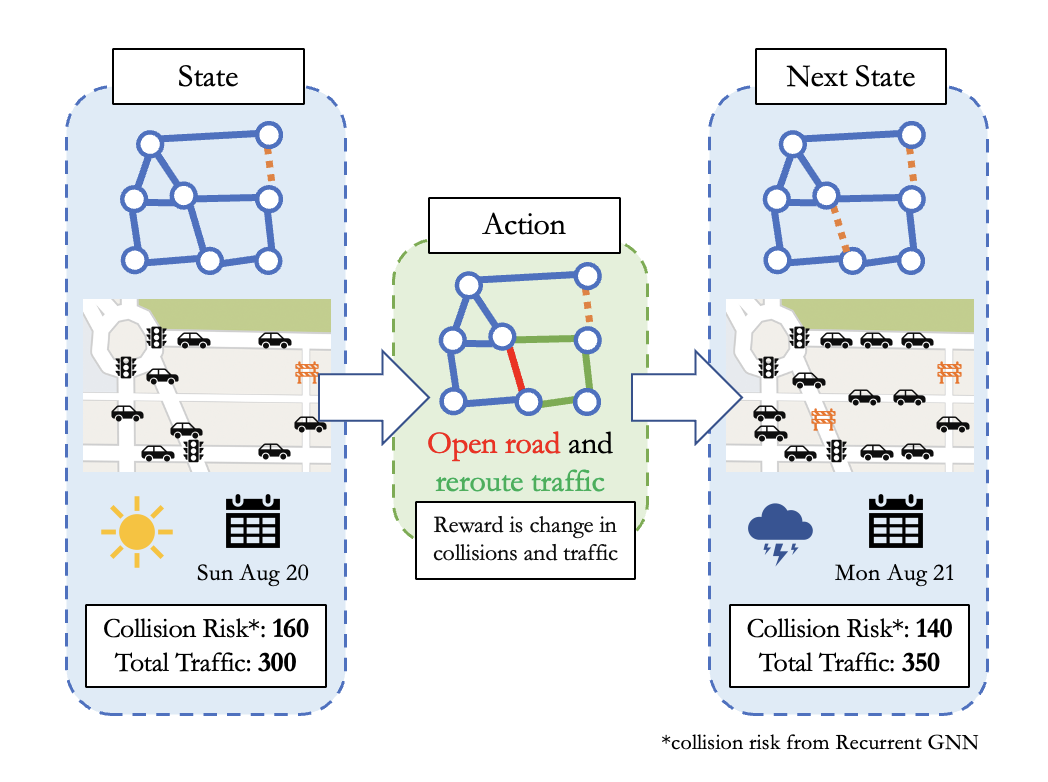}
    \caption{We frame the problem of opening road
    segments in the context of RL.
    A state corresponds to a real historical day where
    we simulated opening road segments.
    An action selects a new road segment to open
    and propagates openings to the next day.}
    \label{fig:rl_problem}
\end{figure}

Figure \ref{fig:rl_problem} shows how we formulate the problem.

\customparagraph{States}
States are representations of the city on a given historical day.
The representation includes weather and traffic, calculated from actual taxi trips.
The representation also includes all infrastructure information
but some road segments have been hypothetically opened (closed to vehicular traffic).
The state carries a list
and updates the traffic pattern in each new day as if
all road segments on the list were opened
(using the approach described below).

\customparagraph{Actions}
From a state, the agent selects a road segment to open.
Opening a road segment requires rerouting all 
cars to alternative routes.
We accomplish this by finding the top $k$
weighted shortest paths in the network where
the road segment is removed and assigning
traffic to each path proportional to its total weight.
The weight of a road segment is the expected time to travel it in optimal conditions:
the product of its posted speed limit and its length.

\customparagraph{Rewards}
For each state, we compute the total collision risk
and the total car density per lane as a measure of traffic.
We compute the density per lane using:
\begin{align*}
\sum_{\textrm{road segments } \ell} \frac{\textrm{cars on $\ell$ per day}}
{\textrm{traffic lanes on } \ell \times \textrm{length of }\ell}.
\end{align*}
However, the collision risk is much more complicated to compute.
The challenge is that the states are hypothetical traffic patterns
on real days so we do not actually know how many collisions
would have occurred.
Our solution is to calculate collision risk using the best model 
for predicting collisions from the first part of our work.
In particular, we compute the predicted collisions
from each state and sum the resulting risk of probability
along each road segment.
We normalize both total collision risk and total traffic using a random day.
In order to compute the reward of an action, we use the sum of the collision risk and traffic in the current state minus the same quantity of the next state.
Then the reward is positive if and only if the next state has reduced collision risk and traffic.
Our general approach is flexible; the sum of collision risk and traffic can easily be reweighted to reflect the priority assigned by domain experts.

The RL agent learns
by sampling trajectories
(sequences of states, actions, and rewards) to find which road segments are best to open.
We investigate 1-month-long trajectories, giving the
agent time to observe the long-term effects of opening
road segments while also experimenting with different strategies.
There are several invalid actions that can
prematurely end a trajectory.
First, opening a road segment is invalid
if there are no cars to reroute
(this can happen because we use a single shortest path
for inferring taxi trips).
Second, opening a road segment is invalid
if there is no other directed path from the starting
intersection to the ending intersection
(this can happen because we limit the road network to Manhattan).

\customparagraph{Local vs. Global Rerouting}
When taking an action,
we consider local rerouting (instead of global rerouting)
because of computational cost.
Our model is equivalent to a setting where drivers determine
their path and then, along the way, find some road segments are opened
and reroute to stay on their chosen path accordingly.
Of course, the more realistic setting is that drivers know which
road segments are opened and incorporate this information
in the path they choose.
Unfortunately, because there are tens of millions of taxi trips in our
data set each month,
we cannot afford to recompute the shortest path
for every action.
If we did, the time it takes to initialize the next state would jump from seconds to hours and Q-learning would be prohibitively slow.

\customparagraph{Q-learning} We solve the RL problem with Q-learning
for two reasons:
First, Q-learning tends to be more sample-efficient than
other RL methods (this is particularly
important because both our
state space and action space are large) \cite{jin2018q}.
Second, Q-learning produces a value which we can interpret
as the expected long-term reward of opening a road segment.
Then the road segment with the largest Q-value corresponds
to the best one to open \textit{while accounting for
endogenous effects and temporal dynamics}.

\begin{figure}[h!]
    \centering \includegraphics[width=1\linewidth]{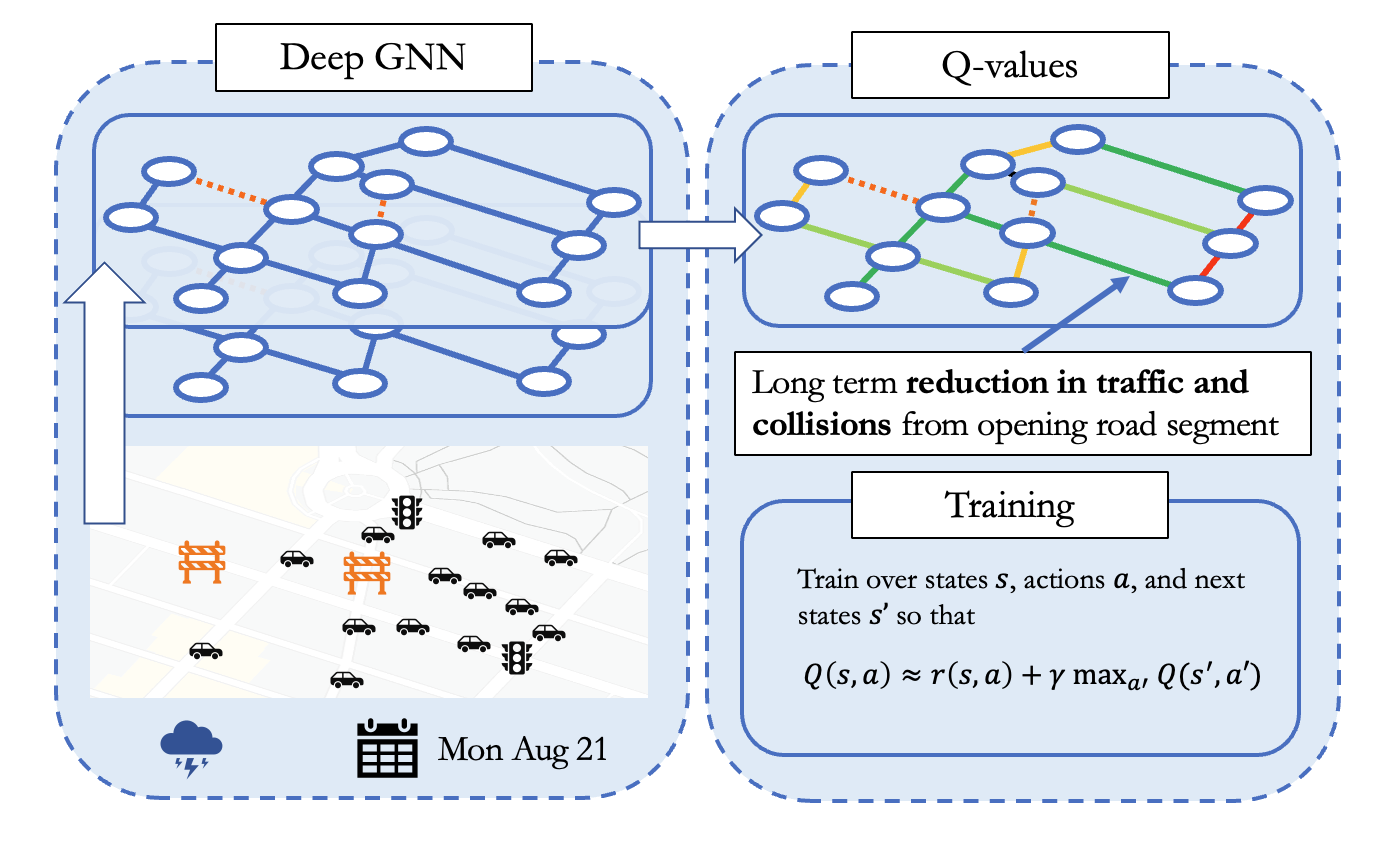}
    \caption{We train the deep Q GNN to output
    the long term expected reduction in traffic
    and collisions from opening each road segment
    in a given day.
    As in general Q-learning, we train so that
    Q-values satisfy the Bellman equation described
    in the background section.}
    \label{fig:q_learning}
    \vspace{-0.3cm}
\end{figure}

Figure \ref{fig:q_learning} describes how we apply Q-learning to the problem of choosing streets.

\begin{figure}
    \centering    \includegraphics[width=1\linewidth]{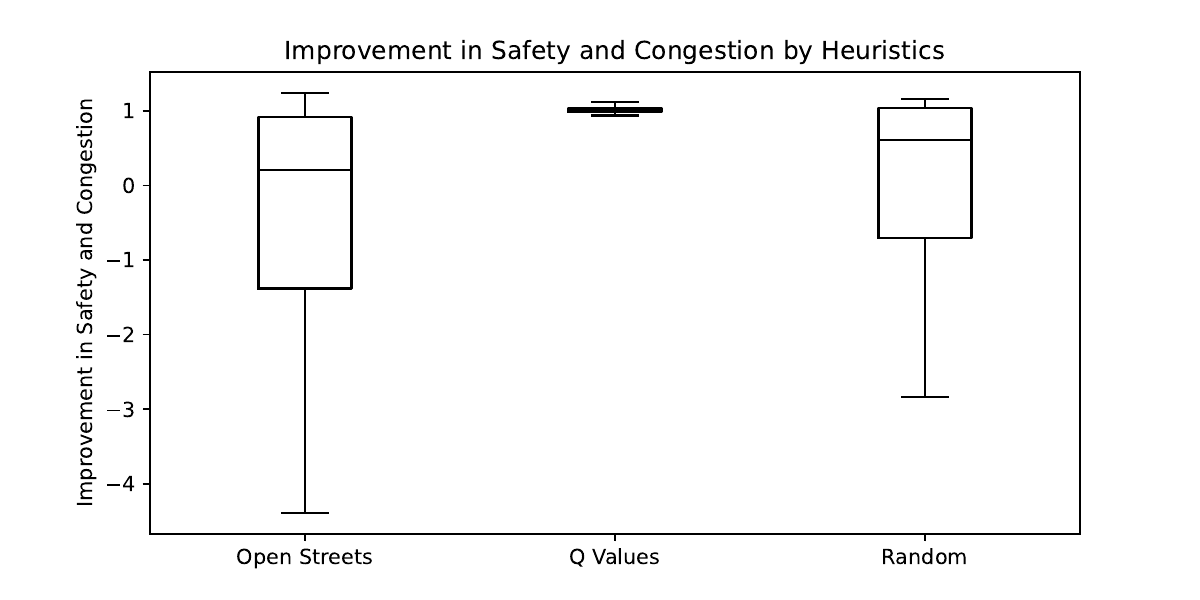}
    \caption{Depicts combined impact of opening a street on safety and collisions. The strategy of choosing streets with the largest Q-value leads to consistently positive reward. In contrast, the streets historically selected by the NYC Open Streets program have high variability and an even worse average impact than a random selection of streets.}
    \label{fig:rl_boxplot}
\end{figure}

\begin{figure}[!ht]
    \includegraphics[width=0.48 \textwidth]{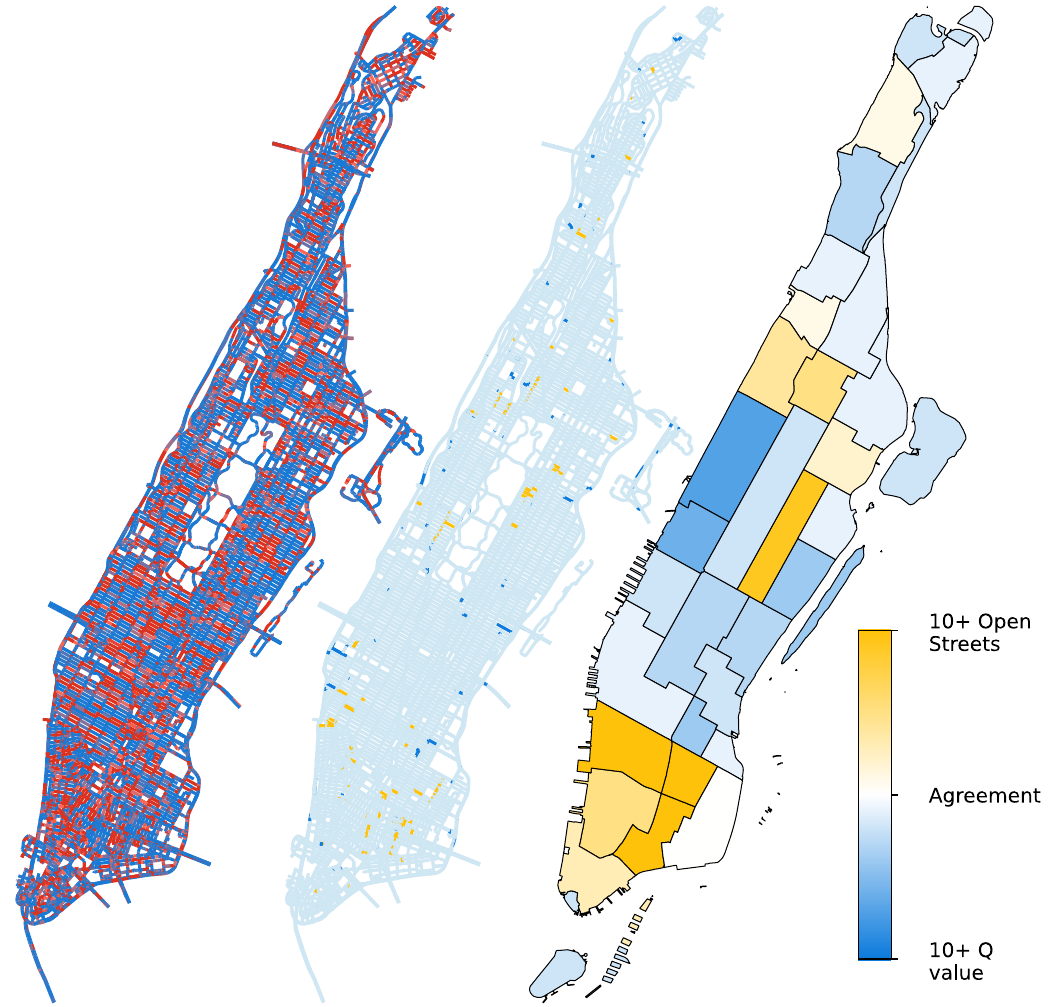}
    \caption{
    Three plots of Manhattan. The left figure plots all 19,391 street segments with their associated Q-values (blue means an expected reduction in collisions and congestion). The middle figure plots the 121 segments in the open streets program (yellow) and the 121 segments with the largest Q-values (blue). The right figure plots neighborhoods colored by the relative prevalence of streets from the open streets initiative (yellow) and streets with the largest Q-values (blue).}
    \label{fig:manhattan}
  \end{figure}

\section{Experiments} \label{sec:experiments}

\label{sec:data}
\customparagraph{Data}
Collision data comes from \textit{NYC Open Data}, which releases cleaned police reports \cite{nyccollisiondata}. Infrastructure data is a road-bed representation of NYC
and contains about a hundred features like road type, traffic direction, and other features for each road segment \cite{nycinfrastructure}. 
Weather data comes from NOAA \cite{weatherdata}, and is geographically coarse (only from a single weather station in Central Park) but updated hourly.
Finally, our taxi trip data comes from NYC's Taxi and Limousine Commission \cite{nyctaxidata}. 
Each trip contains features like the start and end GPS coordinates,
trip duration, and time of trip. 
Because taxi trips stopped being shared with exact start and end locations in 2016, we conducted our experiments on data from 2013, 2014, and 2015.
\customparagraph{Evaluation}
\label{sec:tuning}
We used available implementations of Gaussian NB, XGBoost and LightGBM for collision prediction model baselines. We used (and modified) existing implementations of DSTGCN and Graph WaveNet \cite{yu2021deep, wu2019graph}. We implemented our recurrent GNN models using Pytorch \cite{paszke2019pytorch}. Hyperparameter tuning was done using RayTune \cite{liaw2018tune}, performing random search over a large parameter space with early stopping. 

\paragraph{Computing Resources}
We ran our models through a compute cluster using an A100 Nvidia GPUs with 80GB of RAM. Our non-neural models ran on CPUs.

\section{Experimental Results}


\paragraph{Q-learning provides better long term reward than the open streets initiative status-quo.}
Figure \ref{fig:rl_boxplot} shows a box plot of the reward received from three methods of opening streets.
A positive reward indicates a reduction in congestion and collisions in the days when the streets were opened while a negative reward indicates an increase in congestion and collisions.
Opening the streets with the highest Q-values consistently gives the largest reward. 
Their superior performance makes sense because the q-values were optimized to be a measure of long-term reward.
In contrast, the streets in the NYC open streets initiative were chosen using an application process and a variety of concerns.
Nonetheless,
we find it noteworthy that the streets selected by the open streets program have higher variance and a lower average reward than streets that were randomly selected. 
We believe a principled approach using models like ours can make the open streets initiative have a positive impact on congestion and collisions.
To that end, we envision explicitly modeling modifications and additions to the program as a powerful addition in the toolkit of open street advocates. 

\paragraph{Streets with the highest Q-values are more geographically diverse than those selected by the open streets initiative.}
The middle plot of Figure \ref{fig:manhattan} shows the 121 streets selected by the open streets initiative in yellow and the 121 streets with the highest Q-value in blue.
The streets in the open streets initiative (yellow) are concentrated in Downtown Manhattan and completely absent from Midtown and most of the East Side.
The right plot of Figure \ref{fig:manhattan} confirms the discrepancy.
Most neighborhoods have more streets with large Q-values (light and dark blue) while a handful have many more streets in the open streets initiative (dark yellow).
From these two plots of Manhattan, we find that
the streets with the highest Q-value are more geographically diverse. 
A concern with the open streets initiative is inequity in the neighborhoods that benefit from the program \cite{hazarika2021reclaiming}.
We believe an advantage to using a principled approach is a more equitable geographic distribution.

\paragraph{Most of the worst streets to open are East-West.}
In the left plot of Figure \ref{fig:manhattan}, almost all the streets with the lowest Q-values (dark red) are East-West.
Since Manhattan is optimized for North-South travel with large one-way avenues and synchronized traffic signals, we believe the low Q-values for East-West streets are an emergent property of our model \cite{owen2004green}.

\subsection{Limitations and Future Work}\label{subsec:limitations}
Due to the size of our data (tens of millions of taxi trips per month and 19,391 road segments), we reroute individual taxi trips around opened streets.
This corresponds to the setting where a driver gets to a road and only then learns it is opened, while the more realistic setting corresponds to a driver planning their route with prior knowledge of opened road segments.
Future work could use more compute (or a better technique) to realistically reroute traffic after streets are opened.
We discuss the effective resistance as one such possible technique in the extended version online.

We intentionally focused on NYC to demonstrate a proof-of-concept and integrate feedback from local transportation experts.
We leave applying the approach, and even the networks we trained, to other cities as future work.
We consider the objectives of reducing traffic and collisions.
However, there are more objectives such as pedestrian utilization or tourist interest that could make streets desirable to open.
Future work could integrate other objectives by augmenting the reinforcement learning reward function.
Because collision data is necessarily sparse,
prior work has used cameras and sensors to detect \textit{near-collision} 
events \cite{wang2017vehicle, osman2019prediction}.
Future work could use such data to improve the modeling; however, to the best of our knowledge, the requisite number of cameras and sensors are not available for even a fraction of the segments in Manhattan.

Neural networks are notoriously difficult to
interpret \cite{zhang2021survey}.
This is especially a problem in the high stakes domains of road networks that we applied them to.
We used the integrated gradients method to analyze the feature importance our RGNN but we believe additional interpretability work would benefit models for predicting collisions and opening streets.

\section*{Acknowledgements}
RTW and LR were supported by the National Science Foundation Graduate Research Fellowship under Grant No. DGE-2234660.

\bibliography{aaai24}

\begin{thebibliography}{71}
\providecommand{\natexlab}[1]{#1}

\bibitem[{Abeyratne and Halgamuge(2020)}]{Abeyratne2020ApplyingCity}
Abeyratne, D.; and Halgamuge, M.~N. 2020.
\newblock {Applying Big Data Analytics on Motor Vehicle Collision Predictions in New York City}.
\newblock \emph{Intelligent Data Analysis: From Data Gathering to Data Comprehension}, 219--239.

\bibitem[{Auret and Aldrich(2012)}]{Auret2012InterpretationForests}
Auret, L.; and Aldrich, C. 2012.
\newblock {Interpretation of nonlinear relationships between process variables by use of random forests}.
\newblock \emph{Minerals Engineering}, 35: 27--42.

\bibitem[{Badirli et~al.(2020)Badirli, Liu, Xing, Bhowmik, Doan, and Keerthi}]{badirli2020gradient}
Badirli, S.; Liu, X.; Xing, Z.; Bhowmik, A.; Doan, K.; and Keerthi, S.~S. 2020.
\newblock Gradient boosting neural networks: Grownet.
\newblock \emph{arXiv preprint arXiv:2002.07971}.

\bibitem[{Baloch et~al.(2020)Baloch, Raza, Pathak, Marone, and Ali}]{Baloch2020MachineSeverity}
Baloch, Z.~Q.; Raza, S.~A.; Pathak, R.; Marone, L.; and Ali, A. 2020.
\newblock {Machine Learning Confirms Nonlinear Relationship between Severity of Peripheral Arterial Disease, Functional Limitation and Symptom Severity}.
\newblock \emph{Diagnostics}, 10(8).

\bibitem[{Bao et~al.(2021)Bao, Yang, Zeng, and Shi}]{Bao2021ExploringData}
Bao, J.; Yang, Z.; Zeng, W.; and Shi, X. 2021.
\newblock {Exploring the spatial impacts of human activities on urban traffic crashes using multi-source big data}.
\newblock \emph{Journal of transport geography}, 94: 103118.

\bibitem[{Bertolini(2020)}]{bertolini2020streets}
Bertolini, L. 2020.
\newblock From “streets for traffic” to “streets for people”: can street experiments transform urban mobility?
\newblock \emph{Transport reviews}, 40(6): 734--753.

\bibitem[{Bertsimas, Jaillet, and Martin(2019)}]{Bertsimas2019OnlineApplications}
Bertsimas, D.; Jaillet, P.; and Martin, S. 2019.
\newblock {Online Vehicle Routing: The Edge of Optimization in Large-Scale Applications}.
\newblock \emph{https://doi.org/10.1287/opre.2018.1763}, 67(1): 143--162.

\bibitem[{Boley, Ranjan, and Zhang(2011)}]{boley2011commute}
Boley, D.; Ranjan, G.; and Zhang, Z.-L. 2011.
\newblock Commute times for a directed graph using an asymmetric Laplacian.
\newblock \emph{Linear Algebra and its Applications}, 435(2): 224--242.

\bibitem[{Borisov et~al.(2022)Borisov, Leemann, Se{\ss}ler, Haug, Pawelczyk, and Kasneci}]{borisov2022deep}
Borisov, V.; Leemann, T.; Se{\ss}ler, K.; Haug, J.; Pawelczyk, M.; and Kasneci, G. 2022.
\newblock Deep neural networks and tabular data: A survey.
\newblock \emph{IEEE Transactions on Neural Networks and Learning Systems}.

\bibitem[{Chen et~al.(2012)Chen, Chen, Srinivasan, McKnight, Ewing, and Roe}]{chen2012evaluating}
Chen, L.; Chen, C.; Srinivasan, R.; McKnight, C.~E.; Ewing, R.; and Roe, M. 2012.
\newblock Evaluating the safety effects of bicycle lanes in New York City.
\newblock \emph{American journal of public health}, 102(6): 1120--1127.

\bibitem[{Cheng and Koudas(2019)}]{Cheng2019NonlinearData}
Cheng, Z.; and Koudas, N. 2019.
\newblock {Nonlinear models over normalized data}.
\newblock \emph{Proceedings - International Conference on Data Engineering}, 2019-April: 1574--1577.

\bibitem[{Christiano et~al.(2011)Christiano, Kelner, Madry, Spielman, and Teng}]{christiano2011electrical}
Christiano, P.; Kelner, J.~A.; Madry, A.; Spielman, D.~A.; and Teng, S.-H. 2011.
\newblock Electrical flows, laplacian systems, and faster approximation of maximum flow in undirected graphs.
\newblock In \emph{Proceedings of the forty-third annual ACM symposium on Theory of computing}, 273--282.

\bibitem[{Cohen et~al.(2016)Cohen, Han, Derose, Williamson, Paley, and Batteate}]{cohen2016ciclavia}
Cohen, D.; Han, B.; Derose, K.~P.; Williamson, S.; Paley, A.; and Batteate, C. 2016.
\newblock CicLAvia: Evaluation of participation, physical activity and cost of an open streets event in Los Angeles.
\newblock \emph{Preventive medicine}, 90: 26--33.

\bibitem[{Das, Park, and Sarkar(2023)}]{das2023impact}
Das, S.; Park, E.~S.; and Sarkar, S. 2023.
\newblock Impact of operating speed measures on traffic crashes: Annual and daily level models for rural two-lane and rural multilane roadways.
\newblock \emph{Journal of Transportation Safety \& Security}, 15(6): 584--603.

\bibitem[{Deri, Franchetti, and Moura(2016)}]{Deri2016BigCity}
Deri, J.~A.; Franchetti, F.; and Moura, J. M.~F. 2016.
\newblock {Big data computation of taxi movement in New York City}.
\newblock In \emph{2016 IEEE International Conference on Big Data (Big Data)}, 2616--2625.

\bibitem[{Engstr{\"o}m et~al.(2017)Engstr{\"o}m, Markkula, Victor, and Merat}]{engstrom2017effects}
Engstr{\"o}m, J.; Markkula, G.; Victor, T.; and Merat, N. 2017.
\newblock Effects of cognitive load on driving performance: The cognitive control hypothesis.
\newblock \emph{Human factors}, 59(5): 734--764.

\bibitem[{Furutani et~al.(2020)Furutani, Shibahara, Akiyama, Hato, and Aida}]{furutani2020graph}
Furutani, S.; Shibahara, T.; Akiyama, M.; Hato, K.; and Aida, M. 2020.
\newblock Graph signal processing for directed graphs based on the hermitian laplacian.
\newblock In \emph{Machine Learning and Knowledge Discovery in Databases: European Conference, ECML PKDD 2019, W{\"u}rzburg, Germany, September 16--20, 2019, Proceedings, Part I}, 447--463. Springer.

\bibitem[{Ghasedi~Dizaji et~al.(2017)Ghasedi~Dizaji, Herandi, Deng, Cai, and Huang}]{ghasedi2017deep}
Ghasedi~Dizaji, K.; Herandi, A.; Deng, C.; Cai, W.; and Huang, H. 2017.
\newblock Deep clustering via joint convolutional autoencoder embedding and relative entropy minimization.
\newblock In \emph{Proceedings of the IEEE international conference on computer vision}, 5736--5745.

\bibitem[{Grynbaum(2010)}]{grynbaum2010new}
Grynbaum, M.~M. 2010.
\newblock New york traffic experiment gets permanent run.
\newblock \emph{New York Times}, 11.

\bibitem[{Hazarika(2021)}]{hazarika2021reclaiming}
Hazarika, S. 2021.
\newblock \emph{‘On Reclaiming the Streets for the People’: Understanding Equity in Public Space Planning Strategies Through an Analysis of the Open Streets Program in New York City}.
\newblock Ph.D. thesis, Columbia University.

\bibitem[{INRIX(2022)}]{inrix2022global}
INRIX. 2022.
\newblock Global Traffic Scorecard.
\newblock \emph{INRIX; Kirkland, Washington, USA}.

\bibitem[{Ismaeil, Kholeif, and Abdel-Fattah(2022)}]{Ismaeil2022UsingTaxi}
Ismaeil, H.; Kholeif, S.; and Abdel-Fattah, M.~A. 2022.
\newblock {Using Decision Tree Classification Model to Predict Payment Type in NYC Yellow Taxi}.
\newblock \emph{International Journal of Advanced Computer Science and Applications}, 13(3).

\bibitem[{Jin et~al.(2018)Jin, Allen-Zhu, Bubeck, and Jordan}]{jin2018q}
Jin, C.; Allen-Zhu, Z.; Bubeck, S.; and Jordan, M.~I. 2018.
\newblock Is Q-learning provably efficient?
\newblock \emph{Advances in neural information processing systems}, 31.

\bibitem[{Khosla et~al.(2020)Khosla, Teterwak, Wang, Sarna, Tian, Isola, Maschinot, Liu, and Krishnan}]{khosla2020supervised}
Khosla, P.; Teterwak, P.; Wang, C.; Sarna, A.; Tian, Y.; Isola, P.; Maschinot, A.; Liu, C.; and Krishnan, D. 2020.
\newblock Supervised contrastive learning.
\newblock \emph{Advances in neural information processing systems}, 33: 18661--18673.

\bibitem[{Kuhlberg et~al.(2014)Kuhlberg, Hipp, Eyler, and Chang}]{kuhlberg2014open}
Kuhlberg, J.~A.; Hipp, J.~A.; Eyler, A.; and Chang, G. 2014.
\newblock Open streets initiatives in the United States: closed to traffic, open to physical activity.
\newblock \emph{Journal of physical activity and health}, 11(8): 1468--1474.

\bibitem[{Kwa{\'s}niewska, Rumi{\'n}ski, and Rad(2017)}]{kwasniewska2017deep}
Kwa{\'s}niewska, A.; Rumi{\'n}ski, J.; and Rad, P. 2017.
\newblock Deep features class activation map for thermal face detection and tracking.
\newblock In \emph{2017 10Th international conference on human system interactions (HSI)}, 41--47. IEEE.

\bibitem[{Lainder and Wolfinger(2022)}]{lainder2022forecasting}
Lainder, A.~D.; and Wolfinger, R.~D. 2022.
\newblock Forecasting with gradient boosted trees: augmentation, tuning, and cross-validation strategies: Winning solution to the M5 Uncertainty competition.
\newblock \emph{International Journal of Forecasting}, 38(4): 1426--1433.

\bibitem[{Lana et~al.(2018)Lana, Del~Ser, Velez, and Vlahogianni}]{lana2018road}
Lana, I.; Del~Ser, J.; Velez, M.; and Vlahogianni, E.~I. 2018.
\newblock Road traffic forecasting: Recent advances and new challenges.
\newblock \emph{IEEE Intelligent Transportation Systems Magazine}, 10(2): 93--109.

\bibitem[{Li et~al.(2017)Li, Yu, Shahabi, and Liu}]{li2017diffusion}
Li, Y.; Yu, R.; Shahabi, C.; and Liu, Y. 2017.
\newblock Diffusion convolutional recurrent neural network: Data-driven traffic forecasting.
\newblock \emph{arXiv preprint arXiv:1707.01926}.

\bibitem[{Li et~al.(2013)Li, Wang, Chen, Liu, and Xu}]{li2013evaluation}
Li, Z.; Wang, W.; Chen, R.; Liu, P.; and Xu, C. 2013.
\newblock Evaluation of the impacts of speed variation on freeway traffic collisions in various traffic states.
\newblock \emph{Traffic injury prevention}, 14(8): 861--866.

\bibitem[{Liaw et~al.(2018)Liaw, Liang, Nishihara, Moritz, Gonzalez, and Stoica}]{liaw2018tune}
Liaw, R.; Liang, E.; Nishihara, R.; Moritz, P.; Gonzalez, J.~E.; and Stoica, I. 2018.
\newblock Tune: A research platform for distributed model selection and training.
\newblock \emph{arXiv preprint arXiv:1807.05118}.

\bibitem[{Lin et~al.(2020)Lin, Sun, Bulusu, Dry, and Hernandez}]{lin2020graph}
Lin, C.; Sun, G.~J.; Bulusu, K.~C.; Dry, J.~R.; and Hernandez, M. 2020.
\newblock Graph neural networks including sparse interpretability.
\newblock \emph{arXiv preprint arXiv:2007.00119}.

\bibitem[{Lin et~al.(2021)Lin, Chen, Chiang, and Sharma}]{Lin2021IntelligentApproach}
Lin, D.-J.; Chen, M.-Y.; Chiang, H.-S.; and Sharma, P.~K. 2021.
\newblock {Intelligent traffic accident prediction model for Internet of Vehicles with deep learning approach}.
\newblock \emph{IEEE Transactions on Intelligent Transportation Systems}, 23(3): 2340--2349.

\bibitem[{Liu et~al.(2021)Liu, Singleton, Arribas-Bel, and Chen}]{liu2021identifying}
Liu, Y.; Singleton, A.; Arribas-Bel, D.; and Chen, M. 2021.
\newblock Identifying and understanding road-constrained areas of interest (AOIs) through spatiotemporal taxi GPS data: A case study in New York City.
\newblock \emph{Computers, Environment and Urban Systems}, 86: 101592.

\bibitem[{Luo et~al.(2022)Luo, Shangguan, Yang, Gao, Fang, and Yu}]{Luo2022Spatial-TemporalForecasting}
Luo, A.; Shangguan, B.; Yang, C.; Gao, F.; Fang, Z.; and Yu, D. 2022.
\newblock {Spatial-Temporal Diffusion Convolutional Network: A Novel Framework for Taxi Demand Forecasting}.
\newblock \emph{ISPRS International Journal of Geo-Information}, 11(3): 193.

\bibitem[{Luo et~al.(2019)Luo, Li, Yang, and Zhang}]{luo2019spatiotemporal}
Luo, X.; Li, D.; Yang, Y.; and Zhang, S. 2019.
\newblock Spatiotemporal traffic flow prediction with KNN and LSTM.
\newblock \emph{Journal of Advanced Transportation}, 2019.

\bibitem[{Miglani and Kumar(2019)}]{miglani2019deep}
Miglani, A.; and Kumar, N. 2019.
\newblock Deep learning models for traffic flow prediction in autonomous vehicles: A review, solutions, and challenges.
\newblock \emph{Vehicular Communications}, 20: 100184.

\bibitem[{Moerland et~al.(2023)Moerland, Broekens, Plaat, Jonker et~al.}]{moerland2023model}
Moerland, T.~M.; Broekens, J.; Plaat, A.; Jonker, C.~M.; et~al. 2023.
\newblock Model-based reinforcement learning: A survey.
\newblock \emph{Foundations and Trends{\textregistered} in Machine Learning}, 16(1): 1--118.

\bibitem[{Nilsson et~al.(2017)Nilsson, Ahlstr{\"o}m, Barua, Fors, Lind{\'e}n, Svanberg, Begum, Ahmed, and Anund}]{nilsson2017vehicle}
Nilsson, E.; Ahlstr{\"o}m, C.; Barua, S.; Fors, C.; Lind{\'e}n, P.; Svanberg, B.; Begum, S.; Ahmed, M.~U.; and Anund, A. 2017.
\newblock Vehicle Driver Monitoring: sleepiness and cognitive load.

\bibitem[{NOAA(2022)}]{weatherdata}
NOAA. 2022.
\newblock Daily Summaries Station Detail (Manhattan).
\newblock \emph{National Oceanic and Atmospheric Administration}.

\bibitem[{OpenData(2022)}]{nyccollisiondata}
OpenData, N. 2022.
\newblock Motor Vehicle Collisions.
\newblock \emph{New York City}.

\bibitem[{Osman et~al.(2019)Osman, Hajij, Bakhit, and Ishak}]{osman2019prediction}
Osman, O.~A.; Hajij, M.; Bakhit, P.~R.; and Ishak, S. 2019.
\newblock Prediction of near-crashes from observed vehicle kinematics using machine learning.
\newblock \emph{Transportation Research Record}, 2673(12): 463--473.

\bibitem[{Owen(2004)}]{owen2004green}
Owen, D. 2004.
\newblock Green Manhattan.
\newblock \emph{The New Yorker}, 80(31): 111--23.

\bibitem[{Paszke et~al.(2019)Paszke, Gross, Massa, Lerer, Bradbury, Chanan, Killeen, Lin, Gimelshein, Antiga et~al.}]{paszke2019pytorch}
Paszke, A.; Gross, S.; Massa, F.; Lerer, A.; Bradbury, J.; Chanan, G.; Killeen, T.; Lin, Z.; Gimelshein, N.; Antiga, L.; et~al. 2019.
\newblock Pytorch: An imperative style, high-performance deep learning library.
\newblock \emph{Advances in neural information processing systems}, 32.

\bibitem[{Planning(2022)}]{nycinfrastructure}
Planning, N. 2022.
\newblock LION Single Line Street Base Map.
\newblock \emph{NYC Department of Planning}.

\bibitem[{Ren et~al.(2017)Ren, Song, Wang, Hu, and Lei}]{Ren2017APrediction}
Ren, H.; Song, Y.; Wang, J.; Hu, Y.; and Lei, J. 2017.
\newblock {A Deep Learning Approach to the Citywide Traffic Accident Risk Prediction}.
\newblock \emph{2018 21st International Conference on Intelligent Transportation Systems (ITSC)}, 3346--3351.

\bibitem[{Retallack and Ostendorf(2019)}]{retallack2019current}
Retallack, A.~E.; and Ostendorf, B. 2019.
\newblock Current understanding of the effects of congestion on traffic accidents.
\newblock \emph{International journal of environmental research and public health}, 16(18): 3400.

\bibitem[{Rhoads et~al.(2021)Rhoads, Sol{\'e}-Ribalta, Gonz{\'a}lez, and Borge-Holthoefer}]{rhoads2021sustainable}
Rhoads, D.; Sol{\'e}-Ribalta, A.; Gonz{\'a}lez, M.~C.; and Borge-Holthoefer, J. 2021.
\newblock A sustainable strategy for Open Streets in (post) pandemic cities.
\newblock \emph{Communications Physics}, 4(1): 183.

\bibitem[{Selvaraju et~al.(2017)Selvaraju, Cogswell, Das, Vedantam, Parikh, and Batra}]{selvaraju2017grad}
Selvaraju, R.~R.; Cogswell, M.; Das, A.; Vedantam, R.; Parikh, D.; and Batra, D. 2017.
\newblock Grad-cam: Visual explanations from deep networks via gradient-based localization.
\newblock In \emph{Proceedings of the IEEE international conference on computer vision}, 618--626.

\bibitem[{Seo et~al.(2018)Seo, Defferrard, Vandergheynst, and Bresson}]{seo2018structured}
Seo, Y.; Defferrard, M.; Vandergheynst, P.; and Bresson, X. 2018.
\newblock Structured sequence modeling with graph convolutional recurrent networks.
\newblock In \emph{International conference on neural information processing}, 362--373. Springer.

\bibitem[{Sharples(2014)}]{sharples2014needs}
Sharples, R. 2014.
\newblock \emph{As Needs Must: A Qualitative Study of Motorists’ Habitual Traffic Behaviour in a Situation of Reduced Road Capacity}.
\newblock Ph.D. thesis, University of Technology, Sydney.

\bibitem[{Siemiatycki(2015)}]{siemiatycki2015cost}
Siemiatycki, M. 2015.
\newblock \emph{Cost overruns on infrastructure projects: Patterns, causes, and cures}.
\newblock Institute on Municipal Finance and Governance.

\bibitem[{Singh, Chakraborty, and Manoj(2016)}]{singh2016graph}
Singh, R.; Chakraborty, A.; and Manoj, B. 2016.
\newblock Graph Fourier transform based on directed Laplacian.
\newblock In \emph{2016 International Conference on Signal Processing and Communications (SPCOM)}, 1--5. IEEE.

\bibitem[{Sundararajan, Taly, and Yan(2017)}]{sundararajan2017axiomatic}
Sundararajan, M.; Taly, A.; and Yan, Q. 2017.
\newblock Axiomatic attribution for deep networks.
\newblock In \emph{International conference on machine learning}, 3319--3328. PMLR.

\bibitem[{Sutton and Barto(2018)}]{sutton2018reinforcement}
Sutton, R.~S.; and Barto, A.~G. 2018.
\newblock \emph{Reinforcement learning: An introduction}.
\newblock MIT press.

\bibitem[{Taxi and Commission(2022)}]{nyctaxidata}
Taxi; and Commission, L. 2022.
\newblock NYC Trip Record Data.
\newblock \emph{New York City}.

\bibitem[{Wang and Chan(2017)}]{wang2017vehicle}
Wang, P.; and Chan, C.-Y. 2017.
\newblock Vehicle collision prediction at intersections based on comparison of minimal distance between vehicles and dynamic thresholds.
\newblock \emph{IET Intelligent Transport Systems}, 11(10): 676--684.

\bibitem[{Welling and Kipf(2016)}]{welling2016semi}
Welling, M.; and Kipf, T.~N. 2016.
\newblock Semi-supervised classification with graph convolutional networks.
\newblock In \emph{J. International Conference on Learning Representations (ICLR 2017)}.

\bibitem[{Wu, Wang, and Li(2016)}]{wu2016interpreting}
Wu, F.; Wang, H.; and Li, Z. 2016.
\newblock Interpreting Traffic Dynamics Using Ubiquitous Urban Data.
\newblock In \emph{Proceedings of the 24th ACM SIGSPATIAL International Conference on Advances in Geographic Information Systems}, SIGSPACIAL '16. New York, NY, USA: Association for Computing Machinery.
\newblock ISBN 9781450345897.

\bibitem[{Wu et~al.(2019)Wu, Pan, Long, Jiang, and Zhang}]{wu2019graph}
Wu, Z.; Pan, S.; Long, G.; Jiang, J.; and Zhang, C. 2019.
\newblock Graph wavenet for deep spatial-temporal graph modeling.
\newblock \emph{arXiv preprint arXiv:1906.00121}.

\bibitem[{Xie et~al.(2017)Xie, Ozbay, Kurkcu, and Yang}]{Xie2017AnalysisHotspots}
Xie, K.; Ozbay, K.; Kurkcu, A.; and Yang, H. 2017.
\newblock {Analysis of Traffic Crashes Involving Pedestrians Using Big Data: Investigation of Contributing Factors and Identification of Hotspots}.
\newblock \emph{Risk Analysis}, 37(8): 1459--1476.

\bibitem[{Yang and Gonzales(2017)}]{Yang2017ModelingData}
Yang, C.; and Gonzales, E.~J. 2017.
\newblock {Modeling taxi demand and supply in New York city using large-scale taxi GPS data}.
\newblock In \emph{Seeing cities through big data}, 405--425. Springer.

\bibitem[{Yao et~al.(2018)Yao, Wu, Ke, Tang, Jia, Lu, Gong, Ye, and Li}]{Yao2018DeepPrediction}
Yao, H.; Wu, F.; Ke, J.; Tang, X.; Jia, Y.; Lu, S.; Gong, P.; Ye, J.; and Li, Z. 2018.
\newblock {Deep multi-view spatial-temporal network for taxi demand prediction}.
\newblock In \emph{Proceedings of the AAAI conference on artificial intelligence}, volume~32.

\bibitem[{Yin et~al.(2021)Yin, Wu, Wei, Shen, Qi, and Yin}]{yin2021deep}
Yin, X.; Wu, G.; Wei, J.; Shen, Y.; Qi, H.; and Yin, B. 2021.
\newblock Deep learning on traffic prediction: Methods, analysis, and future directions.
\newblock \emph{IEEE Transactions on Intelligent Transportation Systems}, 23(6): 4927--4943.

\bibitem[{Youn, Jeong, and Gastner(2008)}]{youn2008price}
Youn, H.; Jeong, H.; and Gastner, M. 2008.
\newblock The price of anarchy in transportation networks: efficiency and optimality control', arXiv. org, vol. 0712.

\bibitem[{Yu et~al.(2021)Yu, Du, Hu, Sun, Han, and Lv}]{yu2021deep}
Yu, L.; Du, B.; Hu, X.; Sun, L.; Han, L.; and Lv, W. 2021.
\newblock Deep spatio-temporal graph convolutional network for traffic accident prediction.
\newblock \emph{Neurocomputing}, 423: 135--147.

\bibitem[{Zhang et~al.(2017)Zhang, Sun, Shen, and Zhu}]{Zhang2017AnalyzingData}
Zhang, K.; Sun, D.~J.; Shen, S.; and Zhu, Y. 2017.
\newblock {Analyzing spatiotemporal congestion pattern on urban roads based on taxi GPS data}.
\newblock \emph{Journal of Transport and Land Use}, 10(1): 675--694.

\bibitem[{Zhang et~al.(2021{\natexlab{a}})Zhang, Yu, Guo, Garg, Rodrigues, Hassan, and Guizani}]{zhang2021graph}
Zhang, Q.; Yu, K.; Guo, Z.; Garg, S.; Rodrigues, J.~J.; Hassan, M.~M.; and Guizani, M. 2021{\natexlab{a}}.
\newblock Graph neural network-driven traffic forecasting for the connected internet of vehicles.
\newblock \emph{IEEE Transactions on Network Science and Engineering}, 9(5): 3015--3027.

\bibitem[{Zhang et~al.(2021{\natexlab{b}})Zhang, Ti{\v{n}}o, Leonardis, and Tang}]{zhang2021survey}
Zhang, Y.; Ti{\v{n}}o, P.; Leonardis, A.; and Tang, K. 2021{\natexlab{b}}.
\newblock A survey on neural network interpretability.
\newblock \emph{IEEE Transactions on Emerging Topics in Computational Intelligence}, 5(5): 726--742.

\bibitem[{Zhao et~al.(2019)Zhao, Cheng, Mao, and He}]{Zhao2019ResearchVANET}
Zhao, H.; Cheng, H.; Mao, T.; and He, C. 2019.
\newblock {Research on Traffic Accident Prediction Model Based on Convolutional Neural Networks in VANET}.
\newblock \emph{2019 2nd International Conference on Artificial Intelligence and Big Data (ICAIBD)}.

\bibitem[{Zhong and Sun(2022)}]{zhong2022taxi}
Zhong, S.; and Sun, D.~J. 2022.
\newblock Taxi Driver Speeding: Who, When, Where and How? A Comparative Study Between Shanghai and New York.
\newblock In \emph{Logic-Driven Traffic Big Data Analytics}, 167--182. Springer.

\end{thebibliography}

\clearpage

\appendix

\section{Extended Limitations}\label{subsec:limitations}
We describe several limitations of our work.

\textbf{Global Rerouting}
There are some 10 to 15 million taxi trips per month
and 19391 road segments in Manhattan.
One unfortunate effect is that
computing the shortest path for each trip is time consuming,
even when we group trips by nodes and build predecessor graphs.
As a result, when we open a road, we \textit{locally} reroute drivers.
This corresponds to the setting where a driver gets to a road
and only then learns it is opened.
However, the more realistic is setting would be if we \textit{globally}
reroute drivers.
That is, a driver plans their route knowing which road segments are opened.
Unfortunately, global rerouting would require recomputing 
(basically) all shortest paths in every state of 
the RL trajectory.
This would increase the time to build a state from seconds to hours
and substantially limit the amount of training we give the $Q$ GNN.

\textbf{Real Trip Durations}
Accurately computing the time to travel between two points
is a well studied problem in the transportation literature \cite{lana2018road}.
Unfortunately, there is limited public data since only a few road
segments are monitored by transportation authorities and
access to individual journeys is generally restricted due to privacy concerns.
Instead, recent work has turned to deep learning techniques to
compute trip durations
\cite{miglani2019deep, luo2019spatiotemporal, yin2021deep}.
But, even then, the models need correctly labelled data.
In lieu of another deep learning model with limited labelled data,
we use a simple heuristic for computing the time to travel a road segment:
the product of its length and posted speed limit.
Of course, this is inaccurate especially in settings with substantial
traffic where even large highways can slow to a crawl.

\section{Extended Future Work}\label{subsec:future}
We describe several avenues for future work.

\textbf{Electrical Flow for Directed Graphs}
As described above, we use a simplistic model for inferring routes
that ignores traffic.
Fortunately, there is an elegant model for inferring routes
that takes traffic into account \cite{christiano2011electrical}.
Unfortunately, it only works for undirected graphs.
Consider an undirected graph with nodes $\mathcal{V}$ and edges $\mathcal{E}$.
On this graph, define a flow from some node $s$ to another node $t$.
We call this flow $f:\mathcal{E} \rightarrow \mathbb{R}$ and it
satisfies some constraints 
(basically flow is preserved at each node, with the exception of $s$ and $t$)
Then the energy of the flow is
\begin{align*}
    \sum_{(u,v) \in \mathcal{E}} \frac{f(u,v)^2 }{\textrm{capacity of } (u,v)}.
\end{align*}
It turns out that the flow with minimum energy can be related to
the Laplacian matrix $\mathbf{L}$ of the undirected graph and its
pseudoinverse $\mathbf{L}^\dagger$.
In particular, the potential of each node under the optimal flow
is $\mathbf{L}^\dagger \pmb{\chi}_{st}$
where $\pmb{\chi}_{st} \in \mathbb{R}^{|\mathcal{V}|}$
is a vector of zeros with a 1 in the dimension corresponding to $s$
and a -1 in the dimension corresponding to $t$.
These potentials induce the optimal flow where $f(u,v)$ is the difference
between the the potential on node $u$ and node $v$.
Since the energy can be interpreted as traffic in a network,
the flow minimizes traffic with a \textit{distribution} over edges.
Unfortunately, because this is only defined for undirected graphs, we
cannot use this in Manhattan:
Of the 19391 road segments in our data set,
only 4179 are undirected.
There is some work on building Laplacian matrices for directed
graphs but they either ignore directed information \cite{singh2016graph}
or do not satisfy symmetry
\cite{boley2011commute,furutani2020graph}.
The problem is that the connection of the optimal flow to the Laplacian
requires decomposing the symmetric matrix
into the weighted product of the edge-incidence matrix and its transpose.
As a result, it is unclear to us how generalize 
the relationship between the Laplacian and optimal flow to
directed graphs.

\textbf{Generative Model}
There have been substantial improvements,
and frankly amazing results, in contrastive
learning \cite{khosla2020supervised}
and deep embeddings \cite{ghasedi2017deep}.
One natural idea in the context of our work
is to generate new collisions using
these approaches as an alternative
to predicting collisions with standard
supervised learning techniques.
We leave this problem for future work.
However, we want to add that the imbalanced
nature of collision and non-collision
events pose a problem for effective embeddings.
In particular, it seems that simply
trying to generate realistic collisions
punts the imbalanced nature of the data
down the pipeline.
In addition, even if the model produced
the correct number of collisions, there
would be high variance (since collisions
are so rare) in the RL
states.
This may potentially pose a problem for
effective $Q$ learning.

\textbf{Gradient Boosting 
with Weak Graph Learners}
Gradient boosting techniques like
XGBoost and LightGBM are surprisingly
effective, still outperforming many other
machine learning models in a variety
of data competitions \cite{lainder2022forecasting}.
Traditionally, gradient boosting techniques
use decision trees as weak learners.
However, there has been successful work
replacing the decision trees with shallow neural networks
\cite{badirli2020gradient}.
An interesting idea would be to replace the weak learners
with shallow graph neural networks to capture their
advantages.
Unfortunately, this is beyond the scope of our work.

\textbf{Graph Neural Network Interpretability}
Neural networks are notoriously difficult to
interpret \cite{zhang2021survey}.
This is a problem because of the high stakes
domains, including our application of road networks, 
that we apply them to.
Before neural networks are deployed, we should have
some understanding of why they produce the outputs they do.
In the context of CNNs, there have been several approaches---class
activation mapping \cite{kwasniewska2017deep}
and its gradient successor \cite{selvaraju2017grad}---which successfully
explain why a model outputs what it does.
There is also some work on computing the importance of features
in graph neural networks \cite{lin2020graph}.
We think similar methods should be applied to the graph neural networks
we train before they are implemented in practice.
We used the integrated gradients method to analyze the feature importance of the RGNN but we believe additional interpretability work could benefit the quality of models for predicting collisions and opening streets.


\end{document}